\documentclass[acmsmall]{acmart}
\usepackage{float}
\usepackage{placeins}

\AtBeginDocument{%
  }

\setcopyright{acmlicensed}
\copyrightyear{2018}
\acmYear{2018}
\acmDOI{XXXXXXX.XXXXXXX}

\acmJournal{JACM}
\acmVolume{37}
\acmNumber{4}
\acmArticle{111}
\acmMonth{8}

\begin{document}

\title{Diagnosing and Mitigating Semantic Inconsistencies in Wikidata’s Classification Hierarchy}
\author{Shixiong ZHAO}
\authornote{Both authors contributed equally to this research.}
\email{shixiong@nii.ac.jp}
\orcid{1234-5678-9012}
\author{Hideaki Takeda}
\authornotemark[1]
\email{takeda@nii.ac.jp}
\affiliation{%
  \institution{National Institute of Informatic}
  \city{Chiyoda}
  \state{Tokyo}
  \country{Japan}
}




\begin{abstract}
Wikidata is currently the largest open knowledge graph on the web, encompassing over 120 million entities. It integrates data from various domain-specific databases and imports a substantial amount of content from Wikipedia, while also allowing users to freely edit its content. This openness has positioned Wikidata as a central resource in knowledge graph research and has enabled convenient knowledge access for users worldwide.

However, its relatively loose editorial policy has also led to a degree of taxonomic inconsistency. Building on prior work, this study proposes and applies a novel validation method to confirm the presence of classification errors, over-generalized subclass links, and redundant connections in specific domains of Wikidata. We further introduce a new evaluation criterion for determining whether such issues warrant correction and develop a system that allows users to inspect the taxonomic relationships of arbitrary Wikidata entities—leveraging the platform’s crowdsourced nature to its full potential.

\end{abstract}


\begin{abstract}
Wikidata is currently the largest open knowledge graph on the web, encompassing over 120 million entities. It integrates data from various domain-specific databases and imports a substantial amount of content from Wikipedia, while also allowing users to freely edit its content. This openness has positioned Wikidata as a central resource in knowledge graph research and has enabled convenient knowledge access for users worldwide.

However, its relatively loose editorial policy has also led to a degree of taxonomic inconsistency. Building on prior work, this study proposes and applies a novel validation method to confirm the presence of classification errors, over-generalized subclass links, and redundant connections in specific domains of Wikidata. We further introduce a new evaluation criterion for determining whether such issues warrant correction and develop a system that allows users to inspect the taxonomic relationships of arbitrary Wikidata entities—leveraging the platform’s crowdsourced nature to its full potential.

\end{abstract}
\maketitle

\section{Introduction}

Established in 2012, Wikidata has evolved into one of the internet's most prominent sources of structured information, encompassing over 120 million entities~\cite{Wikidata}. Its success stems from a unique model that combines community contributions with data integration from diverse external sources like Gene Ontology and Wikipedia~\cite{SHENOY2022100679}. This openness has positioned Wikidata as a central resource for knowledge graph research, yet its relatively loose editorial policy introduces significant challenges to its taxonomic integrity and data quality.

While numerous quality issues exist, such as property constraint violations and entity duplication~\cite{SHENOY2022100679}, a particularly fundamental problem lies within its classification hierarchy. The predicates \texttt{P31 (instance of)} and \texttt{P279 (subclass of)} form the taxonomic backbone of the knowledge graph, connecting over 99\% of its entities. However, the misuse and conflation of these two critical relationships have led to what Dadalto et al. (2022) described as a "large-scale conceptual disarray"~\cite{Dadalto2024EvidenceOL}. Their work revealed frequent anti-patterns where entities are simultaneously treated as both instances and classes, undermining the logical consistency of the classification system.

The consequences of such taxonomic inconsistencies are far-reaching. A clear distinction between instances and classes is crucial for the performance of downstream applications, including knowledge graph embedding, where it enables more precise modeling of relational patterns~\cite{lv-etal-2018-differentiating}. Furthermore, structural issues like redundant connections and classification errors can introduce noise that degrades the efficiency and accuracy of reasoning tasks~\cite{Wang2023RiverON}. For a large, multi-domain knowledge graph like Wikidata, maintaining internal structural coherence is therefore essential for its reliability and utility.

While prior work has successfully identified these problems, a comprehensive methodology to diagnose them at scale and guide mitigation efforts is still needed. Simply labeling entities as "erroneous" is often insufficient, as some inconsistencies arise from the multifaceted nature of real-world concepts rather than simple editorial mistakes. This study addresses this gap by proposing a novel, three-stage framework that moves beyond binary error detection to a more nuanced, risk-based assessment of semantic consistency. Our approach combines structural analysis with textual semantic embedding to provide a scalable and multi-faceted diagnostic tool.

The primary contributions of this paper are as follows:
\begin{itemize}
    \item We validate the persistence of structural anti-patterns in a recent Wikidata dump (October 2024), confirming that class-instance confusion remains a significant issue.
    \item We propose a novel, multi-dimensional semantic risk model that quantifies an entity's classification inconsistency based on structural metrics, including parent count, inter-parent distance, and hierarchical depth variance.
    \item We introduce a scalable, LLM-based method for detecting "semantic drift" between an entity and its parent classes by leveraging their textual descriptions, overcoming the computational limitations of graph-based distance metrics for full-graph analysis.
    \item We develop and present a system that integrates our risk metrics, providing a user-facing tool to help Wikidata editors inspect and diagnose semantic inconsistencies in real-time.
\end{itemize}
\section{Related Work}
\subsection{Wikidata Quality}
Wikidata's status as a premier large-scale open knowledge graph is built upon its principles of community control, plurality, and reliance on secondary data\cite{Wikidata}, which have fueled its expansion to over 120 million entities by 2025. However, this scale and openness also introduce significant data quality challenges. A comprehensive study by Shenoy et al. (2021)\cite{SHENOY2022100679} identified a spectrum of quality issues, including property constraint violations, entity duplication, and inconsistent naming conventions. Among these, the conceptual confusion between classes (types) and instances (individuals) stands out as a fundamental and pervasive problem.

Dadalto et al. (2022) \cite{Dadalto2024EvidenceOL}provide compelling evidence of this "large-scale conceptual disarray" within Wikidata's multi-level taxonomies. Through an analysis of anti-patterns involving the P31 (instance of) and P279 (subclass of) relations, their work reveals that a vast number of entities are simultaneously treated as both types and instances. This issue is particularly acute in domains such as biology and biochemistry, where concepts like gene and protein are frequently misrepresented. The authors attribute this problem to several root causes: the inherent ambiguity of natural language constructs like "is-a" (Brachman, 1983\cite{IS-A}), the multi-faceted nature of many real-world concepts, and practical issues such as large-scale automated data imports and the limited adoption of formal classification schemes. These findings underscore the critical need for robust methods to detect and resolve such taxonomic inconsistencies, which is a primary motivation for our current research."

Patel-Schneider \& Doğan (2024) \cite{patelschneider2024classorderdisorderwikidata}conducted a large-scale SPARQL-based analysis of class-order violations in Wikidata, revealing pervasive subclass/superclass misplacements and other hierarchy disorders. They demonstrated that many of these issues can be corrected via manual curation, and provided recommendations for tooling and community practices to improve ontology consistency. This work highlights that taxonomic problems in Wikidata are systemic and repairable.
\subsection{Redundant Connections and KG-based tasks}

In the context of large-scale knowledge graphs such as Wikidata, the issue of edge redundancy presents a nuanced challenge with both detrimental and beneficial aspects. From a data quality and maintenance perspective, redundancy is traditionally viewed as a negative attribute. Foundational database theories, as discussed by Ferrarotti et al. (2009)\cite{Ferrarotti2009FirstOrderTA}, highlight that redundant relations can lead to update anomalies and data inconsistencies. This perspective is echoed in knowledge graph construction, where methodologies often aim to build low-redundant and high-accuracy graphs, treating redundancy as noise that compromises data integrity (Li et al., 2023\cite{low-redun}). Furthermore, formal validation approaches, such as formalizing Wikidata's property constraints using SHACL and SPARQL, implicitly target redundancy, as redundant edges that are not properly maintained can lead to constraint violations and complicate data validation efforts (Ferranti et al., 2024\cite{Ferranti2024FormalizingAV}).

The negative impact of redundancy is particularly pronounced in downstream knowledge graph reasoning tasks. Research by Wang et al. (2023)\cite{Wang2023RiverON} demonstrates that path redundancy in Graph Neural Network (GNN)-based models can increase transformation error entropy, thereby degrading both the efficiency and accuracy of the reasoning process. Their proposed solution involves a mechanism to prune redundant paths during reasoning, underscoring the performance cost associated with such redundancy.

However, a complete removal of all logically redundant edges may not be the optimal strategy. The utility of a knowledge edge is often context-dependent. Yan et al. (2020)\cite{Yan2020LearningCK} propose a model that learns to generate and filter contextualized knowledge structures for specific commonsense reasoning tasks. Their work suggests that an explicitly stated edge, even if logically inferable and thus "redundant", can serve as a powerful and direct signal or a "shortcut" for reasoning. Eliminating such edges globally could remove valuable heuristics and harm the performance on certain tasks. This indicates that a more sophisticated approach than outright deletion is required. The existing literature thus points towards the need for a filtering mechanism, rather than complete removal, to manage redundancy—a perspective that motivates our present work.

\section{Overall Framework}
In our research, we conducted a three-stage analysis of the overall data structure of Wikidata to identify and mitigate semantic ambiguities across its classification system.

In the first stage, we defined a strictly correct reference pattern of \texttt{P31}–\texttt{P279} relationships as a benchmark, and used this structure to compare real-world entity graphs in selected domains. Our results showed that this method effectively identified entities exhibiting composite semantics or classification inconsistencies.

However, we recognized that simply labeling such entities as erroneous and attempting to enforce one-sided corrections is both impractical and theoretically unfounded. Instead, we argue that a more effective approach is to apply a relaxed set of criteria to evaluate whether an entity's classification remains semantically consistent. Once inconsistencies are identified, the entity’s properties should be reorganized across separate semantic dimensions, enabling users to more effectively retrieve context-relevant information.

In the second stage of our work, we began by classifying the uppermost ontology nodes (top-level entities) into conceptual domains that reflect distinct knowledge systems. Using this categorization, we evaluated over four million subclassed entities (i.e., those with \texttt{P279} links) and measured the prevalence of semantic ambiguity among them.

To support this, we constructed a graph-structured representation of Wikidata using the complete set of \texttt{P31} and \texttt{P279} triples. Based on the connectivity patterns among parent classes, intra-graph distances, and hierarchy topology, we defined three dimensions of risk to assess the consistency of each entity’s classification. The resulting scores were integrated into a web-based user interface that enables real-time inspection of the semantic risk associated with any given entity.

In the third stage, we incorporated textual semantics into our model to capture nuances missed by purely structural analysis. We analyzed the semantic alignment between an entity and its parent classes by computing the cosine similarity of their respective textual descriptions using a pre-trained sentence-transformer model. This additional layer allowed us to detect classification errors caused by semantic drift or mismatched contextual meaning.
\section{CME Detection}
\subsection{Methodology}
\begin{figure}[H]
    \centering
    \includegraphics[width=0.8\linewidth]{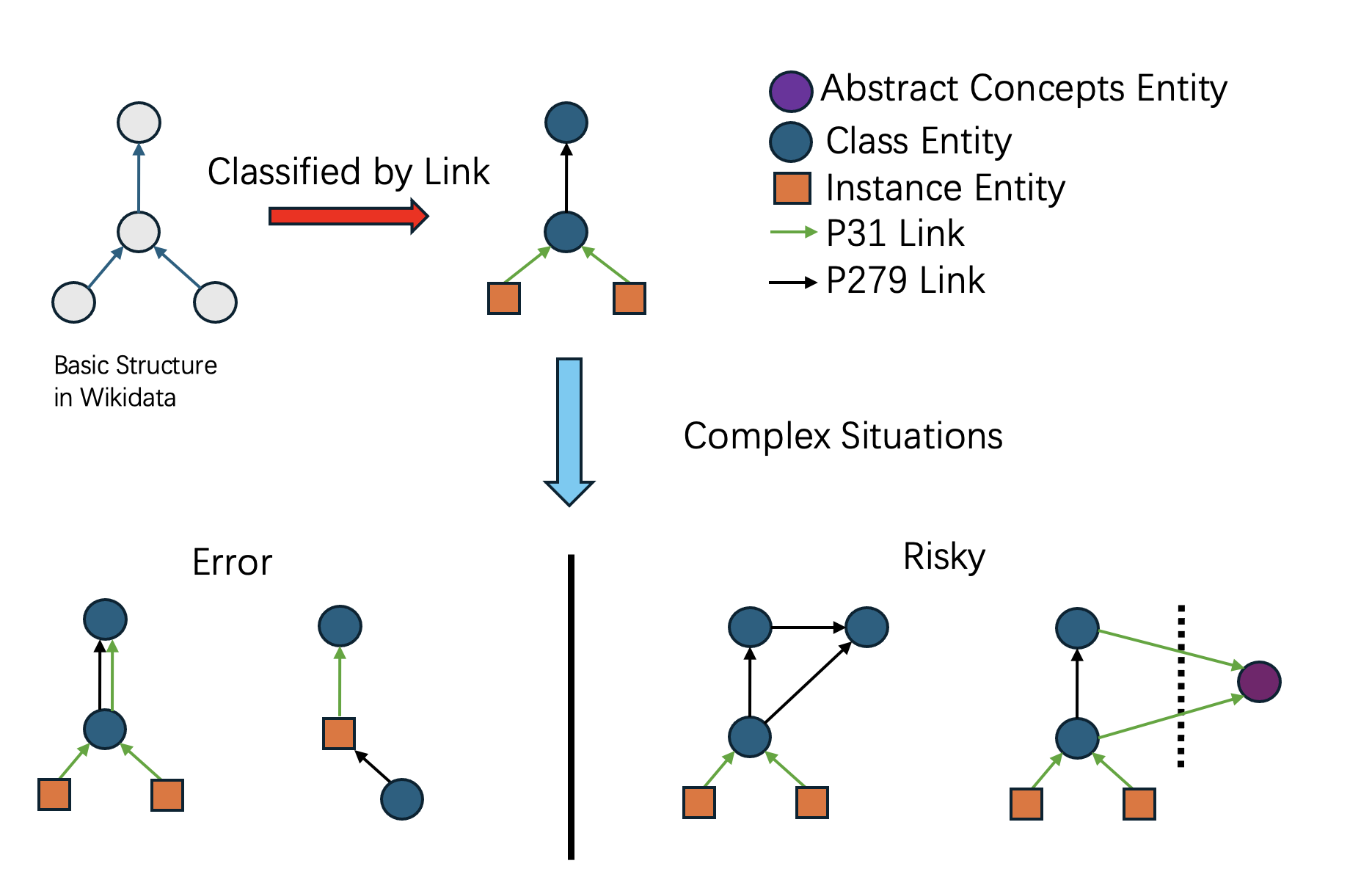}
    \caption{Error Types in P31/P279 Relations}
    \label{fig:Error Types}
\end{figure}

In Wikidata, the mixed use of P31 and P279 can lead to confusion in classification hierarchy and meaning. The Figure \ref{fig:Error Types} shows some typical errors, as well as the link forms we consider correct in this study.
An entity that is linked to a higher level through the P279 relationship should be considered a "class" entity. As a class entity, it can be an instance of the "class" (abstract definition) in Wikidata, but it should not be an instance of other entities with actual meaning. Conversely, if an entity is already linked to a meaningful class entity through P31, it should itself be an instance rather than a subclass, and it should not have P279 relationships from other entities.
The 'complete and correct' patterns depicted in Figure \ref{fig:Error Types} are derived from foundational principles of multi-level modeling theory\cite{Dadalto2024EvidenceOL}, which enforce a strict stratification between instances (0-order entities) and classes (first-order, second-order, etc.). Any deviation from these patterns, such as an entity being both an instance and a subclass, violates these principles.
To detect Composite-Meaning Entities (CMEs) in Wikidata, we designed a graph-based detection pipeline that operates on the July 2024 Wikidata dump containing approximately 100 million entities and limited to P31 (instance of) and P279 (subclass of) triples. Our method consists of the following steps:
\begin{enumerate}
\item \textbf{Graph Segmentation}: We segmented the full P31--P279 graph into weakly-connected components to localize analysis.
\item \textbf{Edge Entity Identification}: Nodes with zero out-degree were used as entry points to initiate traversal.
\item \textbf{Breadth-first Expansion \& Verification}: From each edge node, we expanded the graph breadth-first. A node was flagged if it:
\begin{itemize}
\item Formed a logical contradiction (e.g., simultaneously acted as a class and instance);
\item Introduced taxonomic loops or inherited incompatible roles via mixed use of P31 and P279.
\end{itemize}
\item \textbf{Tagging and Classification}: Flagged entities were recorded with error type tags for follow-up auditing.
\end{enumerate}

\subsection{Results and Evaluation}
We selected six representative knowledge domains and performed five rounds of random sampling. In each round, 100 entities were drawn from each domain, resulting in multiple sample sets for evaluation. For every sampled entity, we examined the graph structure formed through its associated P31 (instance of) and P279 (subclass of) relations, and recorded the number of entities that were ultimately identified as structurally incorrect.

The results confirm observations from prior work: in most knowledge domains, structural inconsistencies caused by the misuse of P31 and P279 are widespread, affecting at least 40\% of sampled entities. In certain domains—particularly those involving highly technical or hierarchically complex knowledge such as genetics or geography—the proportion of incorrect classifications is significantly higher, reaching 80\% to nearly 100\%.The results of our random tests are consistent with previous research findings, which proves that our method in this phase is effective.

\begin{figure}[H]
    \centering
    \includegraphics[width=1\linewidth]{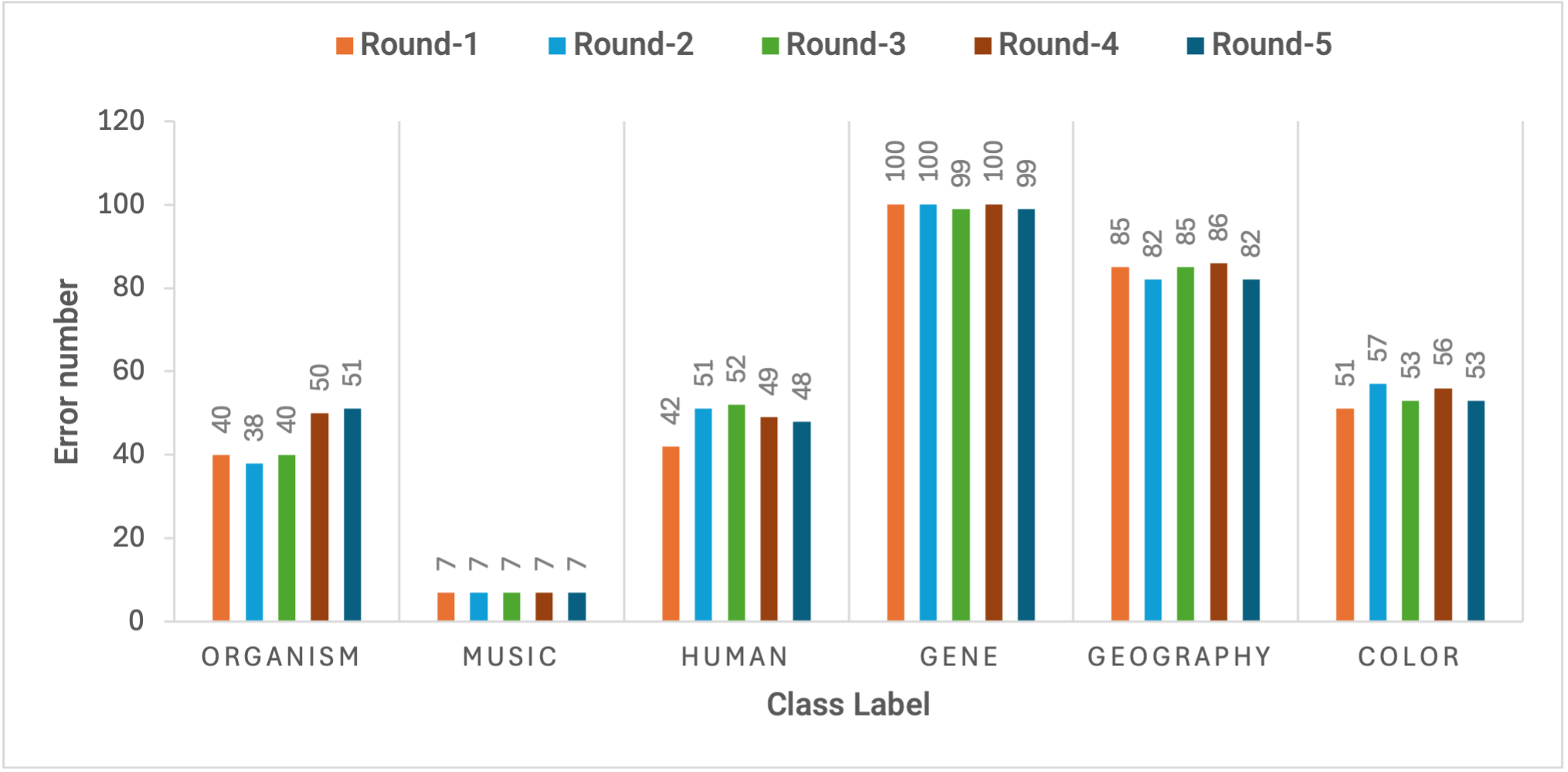}
    \caption{Error Count Across Five Sampling Rounds in Six Knowledge Domains}
    \label{fig:enter-label}
\end{figure}
\FloatBarrier
In order to first gain an understanding of Wikidata as a whole, we analyzed the all entities with respect to P279 (subclass of) and P31(instance of):
There are 4.92 million entities in Wikidata that are used as classes (linked to other entities through the P279 relationship).We define a clean class entity as a class that has exactly one P279 (subclass of) relation and does not participate in any P31 (instance of) relation as a subject, i.e., it does not use P31 to point to any real-world entity class.
After excluding class entities that are not referenced by any instances, we obtain 43,065 entities that satisfy our definition of clean classes. 
For each of these clean classes, we trace their P279 paths upward to identify their corresponding root classes. 
Classes that share the same root are grouped together, forming independent knowledge-domain-specific entity trees.

Through this process, the 43,065 clean classes are organized into 24,793 distinct knowledge domain trees. 
Together, these trees contain approximately 10 million instances, accounting for about 10\% of all Wikidata entities.

\begin{table}[htbp]
\centering
\caption{Overview of Class Hierarchy Characteristics in Wikidata}
\label{tab:wikidata_hierarchy_overview}
\begin{tabular}{lrr}
\toprule
\textbf{Category} & \textbf{Count} & \textbf{Proportion in Wikidata} \\
\midrule

Class Entities & 4,925,023 & 4.3\% \\
Clean Classes Entities(have instances) & 43,065 & 0.03\% \\
Hierarchy Root Clean Classes Entities  & 23,767 & 0.16\% \\
Instances in Clean Classes Entities & 10,458,867 & 9.2\% \\
All Entities in Wikidata Dump & 113,578,686 & 100.0\% \\
\bottomrule
\end{tabular}
\end{table}

It is worth noting that we used very strict rules to filter the data here, which does not indicate that only 10\% of the data in Wikidata is reliable, as the semantic impact of connections has not been considered; rather, excess connections are simply viewed as errors. However, this set of data helps us understand how the ideal Wikidata should be stored and will be used as a reference in subsequent research.

\begin{figure}[H]
    \centering
    \includegraphics[width=1\linewidth]{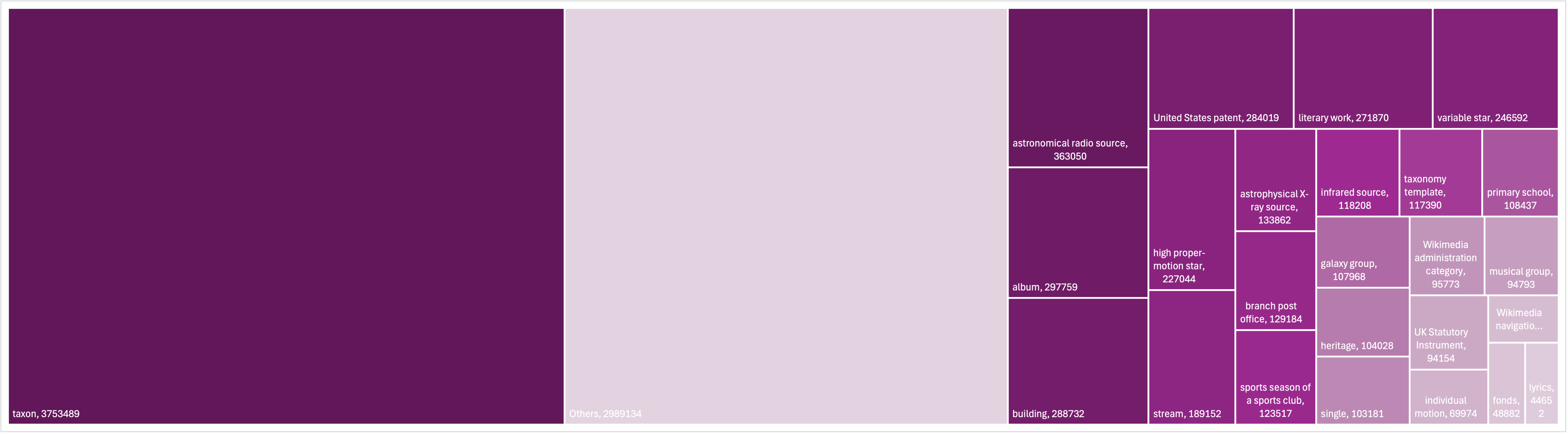}
    \caption{Clean Classes with Instances Number}
    \label{fig:clean_tree_root}
\end{figure}
The figure\ref{fig:clean_tree_root} illustrates the distribution of instances across root class entities in Wikidata. 
Each labeled block represents a root class entity(escape "Others"), the corresponding number indicates the total number of instances subsumed under that class.

The explicitly shown root classes span a diverse range of knowledge domains, including biological taxonomy (e.g., taxon), astronomy and astrophysics (e.g., astronomical radio source, variable star, high proper-motion star), cultural and creative works (e.g., album, literary work, musical group), built environments and institutions (e.g., building, primary school, branch post office), as well as legal and administrative categories (e.g., United States patent, UK Statutory Instrument).
\begin{figure}[H]
    \centering
    \includegraphics[width=1\linewidth]{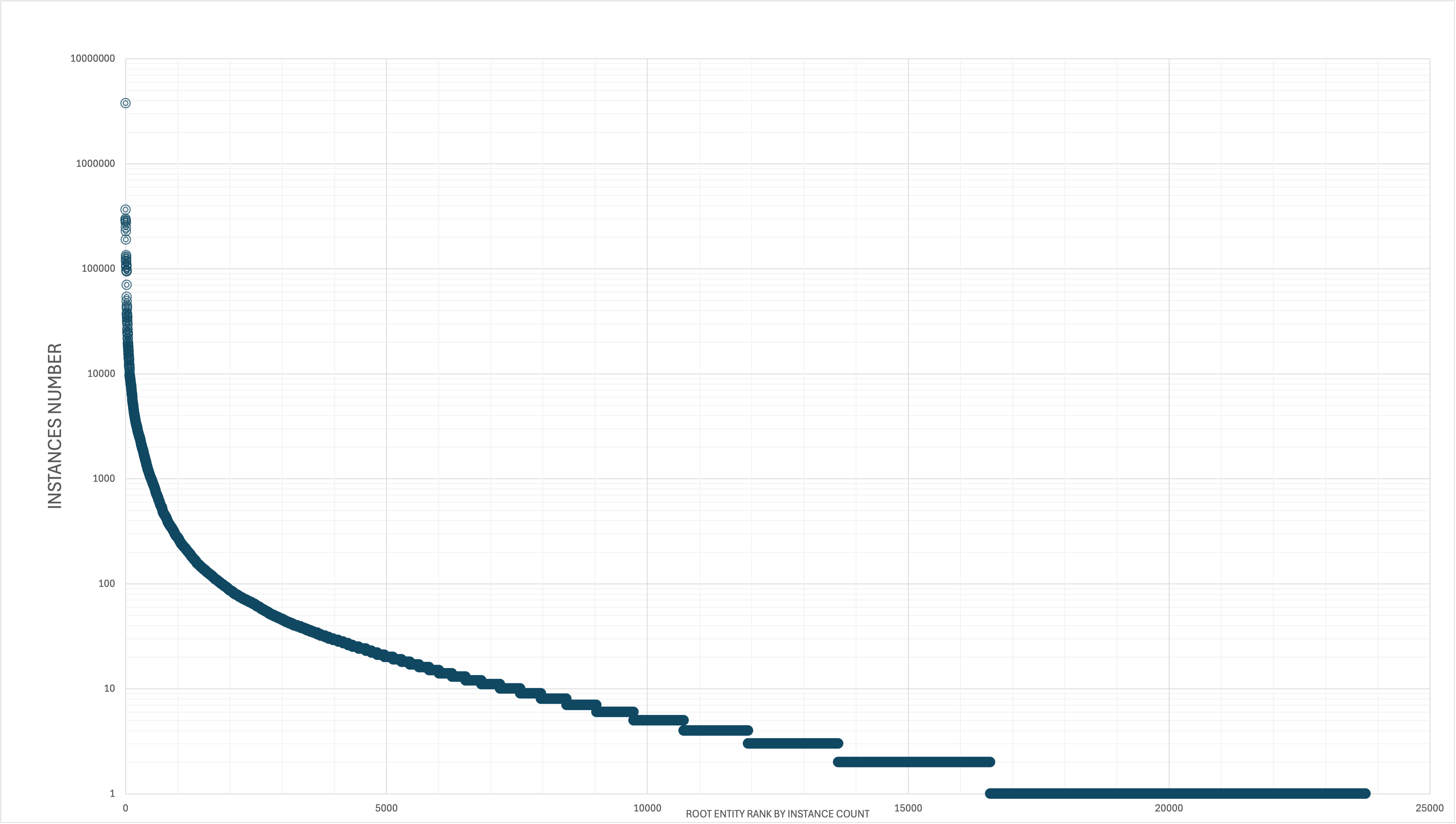}
    \caption{Root Entity Rank by Instances Numbers}
    \label{fig:root_rank_instances_number}
\end{figure}
To quantify the structural distribution hidden within the ``Others'' category, Figure \ref{fig:root_rank_instances_number} plots the instance count against the root entity rank on a semi-logarithmic scale. 
It reveals a distinctive heavy-tailed distribution characteristic in Wikidata.
While the head of the curve exhibits a precipitous drop---confirming that the vast majority of data volume is concentrated in a handful of dominant super-classes---the tail extends significantly with a gradual decay. 
the step-like artifacts observed at the far right of the spectrum (ranks $>10,000$) represent a vast number of niche root classes with minimal instance counts (single digits), highlighting the extreme sparsity and semantic diversity of the taxonomy's fringe.

\section{Entity-Level Semantic Risk Evaluation}

Although the CME detection method in the first stage can effectively identify entities that violate strict taxonomic definitions, it yields only a binary judgment (“problematic” vs. “non-problematic”). We observed that this approach fails to distinguish the severity of inconsistencies. For instance, an entity may be flagged merely due to hierarchical redundancy, whereas another may be flagged for connections to entirely unrelated domains. To enable more fine-grained diagnostics, we designed the second-stage semantic risk assessment metrics, which aim to quantify both the dimensions and the severity of classification inconsistencies.

\begin{figure}[H]
  \centering
  \includegraphics[width=0.8\textwidth]{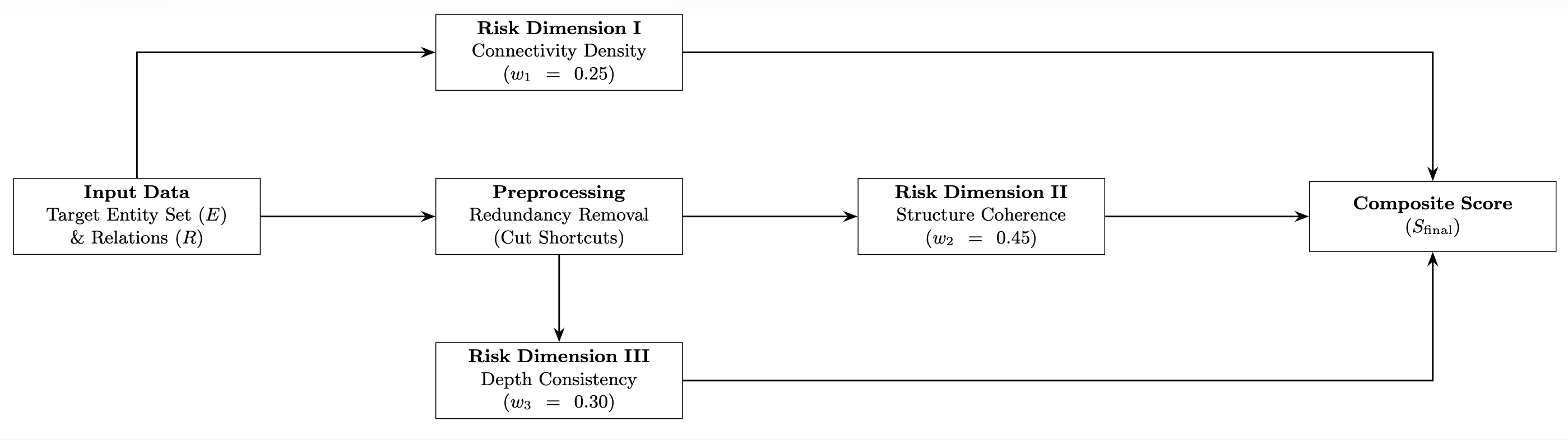}
  \caption{Three Evaluation Dimensions for Entity-Level Semantic Risk.}
  \label{fig:risk_dimensions}
\end{figure}
To assess the degree of semantic ambiguity or multi-meaning risk associated with individual Wikidata entities, we designed a scoring system based on the following three Dimensions:
\begin{enumerate}
  \item \textbf{Connectivity Density (Risk Dimension I)}: \\
  This dimension quantifies the risk of ambiguity derived from the \textbf{raw, unpruned} dataset. Statistics show that the average number of \texttt{P31} and \texttt{P279} relations per entity are approximately 1.068 and 1.001, respectively. An excessively high count suggests that the entity is subject playing multiple conflicting roles or being mapped to overlapping categories.
  
  We calculate the raw connectivity count $N_{raw} = |P31| + |P279|$ and apply Min-Max normalization to map it to the interval $[0, 1]$. 
  \[
  \text{Risk}_{\text{I}}(e) = \frac{N_{raw}(e) - \min(N)}{\max(N) - \min(N)}
  \]
  In our data The Max number of \texttt{P31} and \texttt{P279} is 59, and the Min number is 1.
  A \textbf{higher score} directly indicates greater semantic complexity and higher classification risk.

  \item \textbf{Structural Coherence (Risk Dimension II)}: \\
  This metric measures the extent to which the entity's parent concepts are disjointed or topologically scattered. It is calculated based on the \textbf{cleaned parent set} $U'(e)$ (after removing direct hierarchical redundancies).
  
  We first define the connectivity $\mathrm{conn}_H(c_i, c_j)$ as 1 if the distance $d_G(c_i, c_j) \le 3$, and 0 otherwise. To capture risk, we define the \emph{Structural Coherence} score as the complement of the coherence ratio:
  \[
  \text{Risk}_{\text{II}}(e) = 1 - \frac{
  \sum\limits_{c_i, c_j \in U'(e), i < j} \mathrm{conn}_H(c_i, c_j)
  }{
  \binom{|U'(e)|}{2}
  }.
  \]
  (For $|U'(e)| < 2$, we define $\text{Risk}_{\text{II}}(e) = 0$ as there is no Structural Coherence risk).
  
  Under this formulation, a \textbf{high score (close to 1)} implies that the parent classes are isolated from each other (semantic disjointness), signaling a high risk of concept drift or inconsistent classification. Conversely, a score of 0 indicates a perfectly cohesive cluster.

  \item \textbf{Depth Consistency (Risk Dimension III)}: \\
  This Dimension evaluates the risk associated with inconsistent abstraction levels within the \textbf{cleaned set} $U'(e)$. We calculate the variance of the shortest path lengths to the root node (\texttt{Q35120}):
  \[
  \text{Risk}_{\text{III}}(e) = \mathrm{Var}_{\text{depth}}(e) = \frac{1}{|U'(e)|} \sum_{c_i \in U'(e)} \left( l(c_i) - \bar{l} \right)^2.
  \]
  Since variance measures the spread of data, a \textbf{higher value} inherently indicates that the entity is simultaneously linked to concepts at vastly different granularities (e.g., mixing root-level categories with leaf-level instances). This high disparity reflects taxonomic instability, representing a higher risk of structural error.

\end{enumerate}

The final Composite Risk Score ($S_{risk}$) is computed as a direct weighted sum:
\[
S_{risk}(e) = w_1 \cdot \text{Risk}_{\text{I}}(e) + w_2 \cdot \text{Risk}_{\text{II}}(e) + w_3 \cdot \text{Risk}_{\text{III}}(e)
\]
A higher $S_{risk}$ alerts the system to potentially erroneous or low-quality entity entries.

By aggregating these three metrics, we produce a risk score that reflects the probability of an entity being semantically inconsistent or structurally ambiguous. This score is then surfaced to the user as a diagnostic aid to determine whether further editorial intervention is warranted.

\section{Semantic Drift Detection Methodology}
While the graph-structural risk metrics in the second stage provide in-depth diagnostics, their computational complexity makes them difficult to apply at the scale of the entire Wikidata. To address this scalability challenge, we introduce the third-stage semantic drift detection. This method can be regarded as an efficient proxy for the core ideas of the second stage: by leveraging textual semantic similarity, it approximates the risks arising from connections to semantically distant parent classes , thereby enabling rapid, large-scale scanning of the entire knowledge graph.

This section outlines our methodology for detecting semantic drift in Wikidata entities. Our goal is to identify entities that are semantically inconsistent with their assigned classification, based on both structural signals and embedding-based semantic representations.

This study uses text similarity based on pre-trained language models to approximate semantic consistency, a methodology that is directly inherited from the text enhancement graph embedding paradigm pioneered by KG-BERT (Yao et al., 2019)\cite{yao2019kgbertbertknowledgegraph}. Yao et al. confirmed that treating triples as text sequences input to a Transformer can capture semantic plausibility that cannot be recognized by structural information alone. Unlike KG-BERT, which focuses on link prediction, we adapt this hypothesis to the anomaly detection task: that is, a valid categorical link should exhibit extremely high text implicit in vector space.   

It is worth noting that the term 'semantic drift' used in this paper needs to be distinguished from the term of the same name in the field of ontological evolution. In classical work such as SemaDrift (Stavropoulos et al., 2019)\cite{STAVROPOULOS201987} and OntoDrift (Capobianco et al., 2021)\cite{capobianco2020ontodrift}, semantic drift refers specifically to the evolution of conceptual connotations over time dimensions (i.e., conceptual drift). In contrast, the semantic drift defined in this paper refers to the topological divergence of entity nodes and their taxonomic ancestors in the semantic vector space under a single static snapshot. This divergence measures the 'conceptual stretch' in the classification hierarchy, i.e. the extent to which a particular instance is incorrectly classified into a parent class whose semantic category is too broad or irrelevant.

\subsection{Structural Pre-screening}

We begin with a coarse structural analysis of the Wikidata knowledge graph. Specifically, we extract entity-to-class relations based on the properties \texttt{P31} (\textit{instance of}) and \texttt{P279} (\textit{subclass of}).

To reduce noise, we exclude technical classification nodes such as \texttt{Class} (\texttt{Q16889133}), \texttt{Wikimedia category} (\texttt{Q13442814}), and \texttt{Wikimedia disambiguation page} (\texttt{Q4167410}). The cleaned set of parent relations is denoted as $\mathcal{E}_{\text{clean}}$.

For each entity $e$, we compute:
\begin{itemize}
    \item \textbf{Direct parent count} ($\text{parent\_cnt}_e$): the number of distinct direct parent nodes.
    \item \textbf{Structural depth} ($\text{min\_depth}_e$): the minimum number of hops from $e$ to a pseudo-root node (an entity that is never a child in any relation).
\end{itemize}

We discard entities with only one valid parent, as they provide no meaningful basis for measuring semantic inconsistency. Entities are then grouped into three structural segments:
\begin{itemize}
    \item \textbf{A}: $\text{parent\_cnt} \leq 2$ and $\text{min\_depth} \leq 2$\hfill 
    \item \textbf{C}: $3 \leq \text{parent\_cnt} \leq 6$ and $\text{min\_depth} \leq 2$
    \item \textbf{E}: $\text{parent\_cnt} > 6$
\end{itemize}
We segment entities in this manner to isolate the effects of varying structural complexity. Our hypothesis is that entities with a high number of parent classes (Segment E) may exhibit different patterns of semantic inconsistency compared to those with simpler classification structures. This stratified analysis allows for a more fine-grained understanding of the sources of semantic drift.
\subsection{Text Embedding and Semantic Representation}

We retrieve the textual attributes for each entity---namely its \texttt{label} and \texttt{description}---and concatenate them to construct a semantic representation.

These texts are encoded into dense vectors using the \textbf{Sentence-BERT} model \texttt{all-mpnet-base-v2}, producing normalized embeddings $\vec{e} \in \mathbb{R}^{768}$ for each entity and its parents.
We selected the Sentence-BERT model for its state-of-the-art performance on semantic textual similarity tasks, which is crucial for capturing the holistic meaning of entity descriptions rather than just individual keywords. The all-mpnet-base-v2 variant, in particular, has demonstrated robust performance across various benchmarks and is well-suited for the diverse, general-domain text found in Wikidata.
\subsection{Semantic Drift Computation}

Let $\vec{e}$ be the embedding of a given entity and $\{\vec{p}_1, \dots, \vec{p}_n\}$ the set of its parent embeddings. We define the drift as follows:

\paragraph{Mean Parent Embedding}
We compute the centroid of parent vectors:
\begin{equation}
\vec{p}_{\text{mean}} = \frac{1}{n} \sum_{i=1}^n \vec{p}_i
\end{equation}

\paragraph{Raw Semantic Drift}
We use cosine distance to quantify misalignment:
\begin{equation}
\text{drift}_{\text{raw}} = 1 - \cos(\vec{e}, \vec{p}_{\text{mean}})
\end{equation}

\paragraph{Drift Adjustment}
To account for the number of parents, we apply a logarithmic penalty:
\begin{equation}
\text{drift}_{\text{adj}} = \text{drift}_{\text{raw}} \cdot \log(n + 1)
\end{equation}
We introduce a logarithmic penalty term to adjust the raw drift score. The motivation is that as an entity accumulates more parent classes, the centroid of their embeddings  tends to become more generic. Consequently, a semantically coherent entity might still exhibit a high raw drift simply due to the semantic breadth of its parents. The logarithm function moderately amplifies this effect, penalizing a high parent count while preventing excessive penalization for entities with a very large number of parents.

\begin{figure}[H]
    \centering
    \includegraphics[width=1\linewidth]{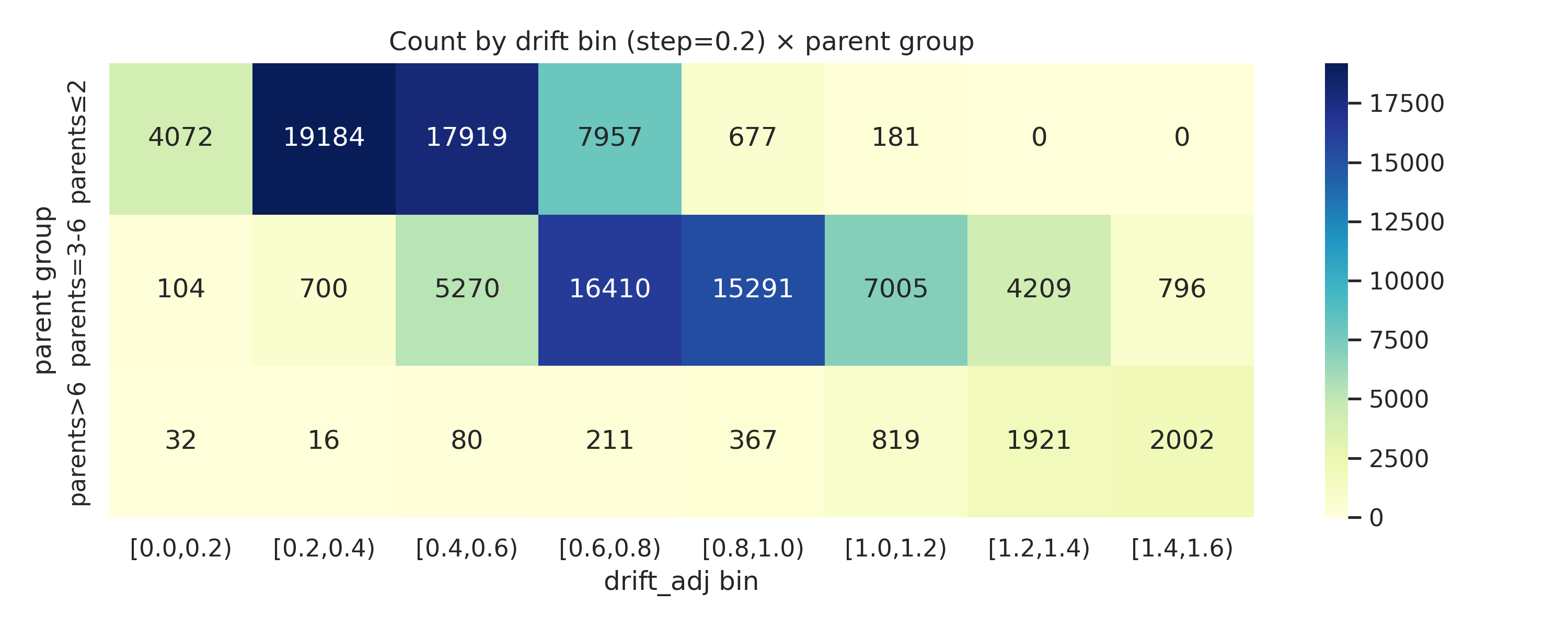}
    \caption{Heatmap of Entity Counts Distributed by Adjusted Drift Score and Parent Group}
    \label{fig:heatmap_count}
\end{figure}

\paragraph{Thresholding}
Entities with $\text{drift}_{\text{adj}} \geq 0.60$ are flagged as semantically inconsistent.
The selection of this threshold is empirically justified by the global distribution of drift scores across different structural complexities, as detailed in Figure\ref{fig:heatmap_count}.
Figure\ref{fig:heatmap_count} shows that the distribution of semantic drift varies substantially with an entity’s structural complexity, measured by the number of its parent classes.

Entities with two or fewer parents, which constitute the vast majority of the dataset, are highly concentrated in low-drift regions (adjusted drift < 0.6). For this group, the entity count declines sharply beyond the [0.4, 0.6) bin, indicating that higher drift values are statistically rare and are more likely to correspond to genuine semantic inconsistencies. In contrast, entities with more complex structures (3–6 parents and more than 6 parents) exhibit distributions that are clearly shifted toward higher drift scores, with their frequencies peaking in the [0.6, 1.0) range.

The value 0.6 therefore emerges as a critical inflection point. It effectively isolates outliers among structurally simple entities while remaining sensitive to the inherently higher drift levels associated with structurally complex entities. As such, this threshold provides a reasonable and data-driven criterion for flagging potential semantic drift across the knowledge g

\subsection{Case Study and Results}

To provide a more concrete understanding of what these drift scores represent, Table\ref{tab:drift_examples} presents a qualitative analysis of representative entities sampled from low, medium, and high-drift regions of the distribution.

As illustrated, low-drift entities such as Endoglycosylceramidase (Q5376341) exhibit strong semantic cohesion among their parent classes, which are all tightly related concepts within the domain of biochemistry. This results in a very low adjusted drift score of 0.196. In contrast, entities with high drift scores demonstrate significant semantic divergence. For example, Nishi-Wakamatsu Station (Q7040449), with a drift score of 1.482, is simultaneously classified under multiple parent classes that describe disparate aspects such as its function (e.g., junction station), physical structure (e.g., over-track railway station), and conceptual type (e.g., human settlement). This conflation of different semantic dimensions in its classification is precisely what our metric is designed to capture as high semantic drift. A moderate-drift case like mining of metal ores (Q16638398) reveals another common issue: hierarchical redundancy, where its parent classes (mining, extractive industry, industry) are semantically related but exist at different levels of abstraction.
\begin{table*}[ht]
\centering
\caption{Representative Examples of Semantic Drift Scores}
\label{tab:drift_examples}
\resizebox{\textwidth}{!}{%
\begin{tabular}{@{}lllll@{}}
\toprule
\textbf{Drift Level} & \textbf{Example Entity (QID)} & \textbf{Adj. Drift} & \textbf{Parent Classes} & \textbf{Qualitative Analysis} \\ \midrule
Low Drift & Endoglycosylceramidase (Q5376341) & 0.196 & 
\begin{tabular}[t]{@{}l@{}}
- group or class of enzymes \\ 
- glycoside hydrolase superfamily \\ 
- hydrolase, hydrolyzing...
\end{tabular} & 
\begin{tabular}[t]{@{}p{5cm}@{}}
\textbf{High Semantic Cohesion:} All parent classes precisely describe the entity's biochemical role. The semantics are highly focused and consistent.
\end{tabular} \\ \midrule
Moderate Drift & \begin{tabular}[t]{@{}l@{}}mining of metal ores \\(Q16638398)\end{tabular} & 0.681 & 
\begin{tabular}[t]{@{}l@{}}
- mining \\ 
- extractive industry \\ 
- industry
\end{tabular} & 
\begin{tabular}[t]{@{}p{5cm}@{}}
\textbf{Hierarchical Redundancy:} Parent classes are semantically related but exist at different levels of abstraction (`mining` is a type of `industry`). This creates moderate drift.
\end{tabular} \\ \midrule
High Drift & \begin{tabular}[t]{@{}l@{}}Nishi-Wakamatsu Station \\(Q7040449)\end{tabular} & 1.482 & 
\begin{tabular}[t]{@{}l@{}}
- railway station \\ 
- junction station \\ 
- over-track railway station \\ 
- last station \\ 
- \textbf{human settlement} \\ 
- ... (and 5 more)
\end{tabular} & 
\begin{tabular}[t]{@{}p{5cm}@{}}
\textbf{High Semantic Divergence:} Parent classes describe disparate aspects: function (junction), structure (over-track), and conceptual type (human settlement), leading to severe inconsistency.
\end{tabular} \\ \bottomrule
\end{tabular}%
}
\end{table*}

These examples empirically validate that our proposed metric effectively quantifies the degree of semantic consistency in an entity's classification hierarchy, capturing issues ranging from hierarchical redundancy to severe semantic divergence.
\begin{figure}[H]
    \centering
    \includegraphics[width=1\linewidth]{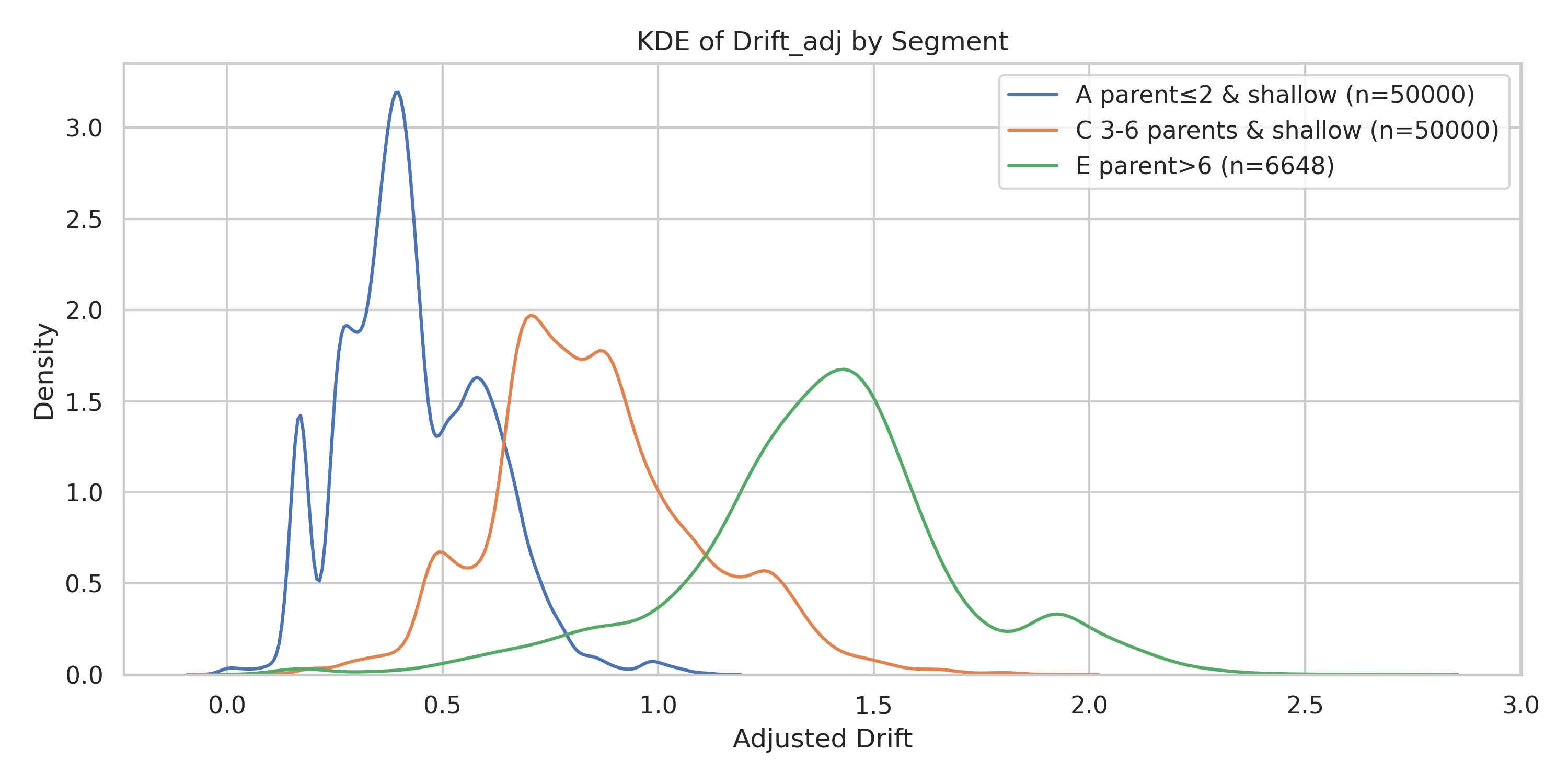}
    \caption{Semantic Drift in Different Group of Entities}
    \label{fig:semantic_drift_kde}
\end{figure}
Figure \ref{fig:semantic_drift_kde} presents a Kernel Density Estimation (KDE) plot that reveals a distinct correlation between structural complexity and semantic drift. The distribution of drift scores exhibits a clear monotonic shift to the right as the number of parent classes increases:

\begin{itemize}
    \item \textbf{High Coherence in Simple Structures (Group A):} The blue curve, representing entities with $\le 2$ parents, is characterized by a sharp, narrow distribution concentrated in the low-drift region (peak $\approx 0.35$). The density drops precipitously after 0.60, confirming that structurally simple entities typically maintain high semantic cohesion. This sharp drop-off empirically validates our selection of 0.60 as a robust threshold for detecting anomalies in the majority of Wikidata entities.
    
    \item \textbf{Inherent Divergence in Complex Structures (Group E):} Conversely, the green curve (entities with $>6$ parents) shows a significantly broader and flatter distribution with a peak around 1.5. This "long-tail" behavior indicates that entities with heavy distinct inheritance chains are inherently more prone to semantic divergence. The high drift scores here reflect both the cumulative semantic distance of diverse parent concepts and the logarithmic penalty term in our metric.
    
    \item \textbf{Transitional Behavior (Group C):} The orange curve serves as a bridge, showing that moderate structural complexity ($3-6$ parents) leads to a predictable increase in semantic ambiguity, centered around the 0.6-0.9 range.
\end{itemize}

\section{Interface-Supported}

The system includes an interactive web-based interface built with Streamlit, designed to support multilingual analysis and visualization of Wikidata classification metrics. The interface allows users to input entity identifiers (QIDs), configure graph traversal parameters, and view diagnostic results in a structured and interpretable format.

The interface is divided into three major functional areas:

\begin{itemize}
    \item Control Panel (Left Sidebar):
Provides user controls including multilingual switching (Chinese, Japanese, English), CSV directory selection, QID input, and maximum path configuration for redundancy detection.
    \item Entity Summary and Metric Dashboard (Center Panel):
Displays the selected entity’s name, description, and overall risk score. It also presents a concise “Strengths \& Potential Issues” section and a weighted breakdown of metric components such as Connection Count, Structural Coherence, Depth Variance, and Instance-Class Alignment.
    \item Semantic Visualization (Right Panel):
Includes metric summary tables, node-level redundancy diagnostics, and semantic similarity heatmaps derived from text embeddings (sentence-transformers/all-mpnet-base-v2). This area highlights hierarchical stability and semantic divergence through color-coded visual cues.
\end{itemize}

\begin{figure}[H]
    \centering
    \includegraphics[width=1\linewidth]{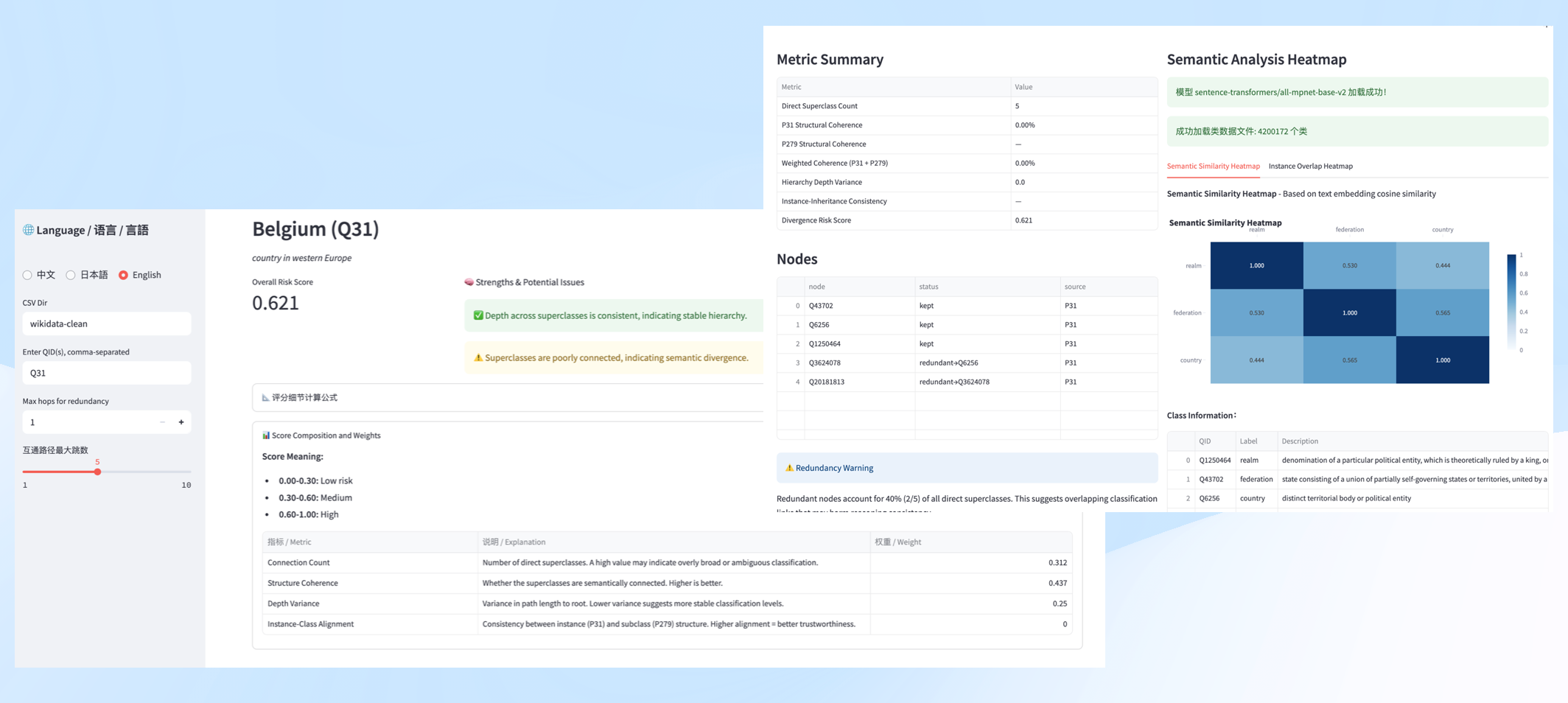}
    \caption{Interface for Entity Check System}
    \label{fig:interface}
\end{figure}

\section{Discussion}

\subsection{Redefining Semantic Inconsistency: From "Error" to "Risk"}
A primary contribution of this research is the reframing of semantic inconsistency not as a binary error to be corrected, but as a quantifiable risk to be managed. Traditional quality assessments often apply a dichotomous correct/incorrect label, a paradigm that proves insufficient for a complex, community-curated knowledge graph like Wikidata. As evidenced by Dadalto et al. (2022) \cite{Dadalto2024EvidenceOL}, many taxonomic anti-patterns arise from the inherently multifaceted nature of real-world concepts, suggesting that simplistic, one-sided corrections are often inappropriate.

Our "semantic risk" framework offers a more nuanced alternative. It acknowledges that inconsistencies exist but quantifies them as a spectrum of risk. This approach respects Wikidata's core principle of "plurality of facts" \cite{Wikidata} while providing actionable insights for downstream applications. For example, an entity with a high-risk score might be flagged for careful handling in high-precision tasks like commonsense reasoning, yet its connections could remain valuable in others. This perspective also informs the debate on redundant edges. Rather than wholesale removal, our risk model suggests a filtering mechanism: high-risk redundant links (e.g., those connecting to semantically distant parents) should be prioritized for review, while low-risk ones (e.g., hierarchical shortcuts) could be preserved to serve as useful heuristics, as suggested by Yan et al. (2020) \cite{Yan2020LearningCK}.

\subsection{The Interplay and Tension Between Structure and Semantics}
Our three-stage methodology highlights the complementary yet sometimes tense relationship between structural and semantic analysis in knowledge graphs. The initial stages, relying on the graph's topology (P31/P279 links), are effective at identifying clear logical contradictions and establishing a baseline for inconsistency, as demonstrated in our CME detection pipeline. However, this structural approach has inherent limitations: it is computationally expensive at scale and cannot capture subtle semantic drifts where the taxonomy is structurally plausible but conceptually flawed.

The introduction of text-based semantic drift detection in our third stage directly addresses these limitations. The case of Nishi-Wakamatsu Station (Q7040449) is illustrative: while its structural connection to human settlement might be short, the semantic chasm between these concepts is vast, a fact our embedding-based model correctly identifies. This underscores a critical tension: which signal should be trusted more when structure and semantics conflict? When an entity is structurally coherent but exhibits high textual drift, it raises profound questions about the nature of community-curated knowledge—whether to prioritize the explicit, user-defined hierarchy or the implicit semantics embedded in textual descriptions. This tension is not a weakness of the model but a reflection of the complex reality of Wikidata, pointing toward future research in hybrid validation models.

\subsection{Generality and Limitations of the Methodology}
The proposed framework, particularly the semantic drift detection stage, is designed for scalability and generality. The use of pre-trained language models makes it largely language-agnostic (though tested here on English) and applicable to any domain within Wikidata or other large-scale knowledge graphs.

However, we acknowledge several limitations. First, the efficacy of our third-stage method is contingent upon the quality and availability of textual descriptions for entities. Entities with sparse or poorly written descriptions may yield unreliable drift scores. Second, our analysis is currently focused on the foundational P31 and P279 relations. While these form the taxonomic backbone, other properties like P361 (part of) can also imply hierarchical relationships, which our current model does not consider. Future work could expand the scope of analysis to a broader set of relational types. Finally, the selection of the 0.60 threshold for high drift, while empirically justified by the data distribution (Figure 3), remains a hyperparameter. Its optimal value may vary across different knowledge domains or for different downstream tasks, suggesting a potential for adaptive, context-aware thresholding in future iterations.

\subsection{Practical Implications for Wikidata and Downstream Applications}
The outcomes of this research have significant practical implications for both the Wikidata community and the developers of knowledge-graph-driven applications.

For the Wikidata Community: The entity-level risk scores and the user interface developed in our second stage can serve as a powerful diagnostic tool for editors. Rather than merely flagging errors, our system provides a multi-dimensional view of why an entity's classification might be problematic. This empowers editors to make more informed decisions, facilitating a more targeted and effective crowdsourced curation process and helping to improve data quality at the source.

For Downstream Applications: The semantic risk score can be directly integrated into various applications to enhance their robustness.

Knowledge Graph Refinement: The scores can be used to prioritize entities for automated or semi-automated cleaning tasks.

KG Embedding: Models can leverage the risk score to weight the importance of taxonomic links during the embedding process, assigning lower weights to high-risk, inconsistent parent relationships.

Retrieval-Augmented Generation (RAG): In RAG systems that retrieve facts from Wikidata, the risk score can function as a reliability signal. Facts associated with high-risk entities can be down-weighted or trigger further verification steps, leading to more accurate and trustworthy generated outputs.
\section{Conclusion}
In this study, we conducted a comprehensive investigation into the structural and semantic inconsistencies within Wikidata’s classification hierarchy, with a focus on the instance of (P31) and subclass of (P279) predicates that define the taxonomic backbone of the graph.

We began by reproducing and extending prior findings through a structure-based comparison method, using ideal graph templates to identify entities that deviate from taxonomic norms. Our analysis, based on the 2024 Wikidata dump, confirmed the continued presence of classification errors—including redundant edges and instance-class confusion—across multiple knowledge domains.

In the second phase, we introduced a set of semantic risk evaluation metrics to quantify the degree of inconsistency at the entity level. These metrics—covering type/class count, inter-parent distance, and parent-level variance—allowed for a more fine-grained diagnosis of semantic ambiguity. To support practical usage, we developed a system that enables users, particularly Wikidata editors, to inspect and visualize classification risks for any given entity or entity list.

To address scalability challenges associated with graph-based analysis, the third phase of our research leveraged pretrained language models to estimate semantic divergence based on textual descriptions. By embedding and comparing entity and parent class descriptions, we efficiently detected semantic drift across millions of entities, enabling full-graph analysis within a short timeframe.

Our findings highlight that while structural violations can pose risks to semantic integrity, not all deviations from taxonomic ideals are inherently harmful. In certain contexts, particularly those involving knowledge graph embedding or question answering, select redundant links may serve as useful shortcuts. These results suggest that future work on knowledge graph curation—especially in crowdsourced settings like Wikidata—should balance taxonomic rigor with task-specific utility.

\section{Future Works}
To ensure the rigor of the research and the effectiveness of the proposed methods, future experimental validation should focus on the following aspects:

\subsection{Expanded Validation Datasets}
The proposed methods should be tested on more diverse subsets of Wikidata, extending beyond the knowledge domains currently selected in this study, to validate their generalizability and robustness across various scenarios. Experimental design could consider incorporating subsets that differ in scale, thematic coverage, and community editing activity levels.

\subsection{Benchmark Comparison Analysis}
It is recommended to conduct systematic benchmark comparisons of the proposed methods against existing Wikidata cleansing tools, quality assessment frameworks, or classification refinement techniques. For instance, comparisons could be made with the taxonomy refinement methods described in the WiKC framework (Peng et al., 2024) or the classification structure cleaning strategies employed in YAGO 4.5 (Pulido-Garcés et al., 2023), depending on their public availability and reproducibility. Such comparisons would facilitate an objective assessment of the relative advantages and potential limitations of the methods developed in this research.

\subsection{Multi-dimensional Evaluation Metrics}
The evaluation framework should extend beyond traditional metrics such as precision, recall, and F1-score for error detection. It is advisable to incorporate broader evaluation dimensions:
\begin{itemize}
    \item \textbf{Impact on Knowledge Graph Coherence}: Assessing improvements in the logical consistency of the Wikidata classification system, including eliminating cyclical structures and redundant relationships.
    \item \textbf{Potential Impact on Downstream Application Performance}: Evaluating how the refined Wikidata classification system enhances the accuracy and efficiency of downstream tasks such as question answering, entity linking, and knowledge recommendation.
    \item \textbf{Computational Efficiency and Scalability}: Measuring computational resource consumption, execution time, and scalability of the proposed methods when processing large-scale Wikidata data.
    \item \textbf{Manual Evaluation and In-depth Case Studies}: Conducting detailed manual sampling and evaluation of classification inconsistencies detected automatically, or refinement suggestions proposed by the methods, to verify their accuracy and practical applicability. Further, in-depth analysis of specific complex or edge cases can reveal strengths and weaknesses of the methods and directions for future improvements.
\end{itemize}

\bibliographystyle{plain}
\bibliography{acmsmall}

@inproceedings{lv-etal-2018-differentiating,
    title = "Differentiating Concepts and Instances for Knowledge Graph Embedding",
    author = "Lv, Xin  and
      Hou, Lei  and
      Li, Juanzi  and
      Liu, Zhiyuan",
    editor = "Riloff, Ellen  and
      Chiang, David  and
      Hockenmaier, Julia  and
      Tsujii, Jun{'}ichi",
    booktitle = "Proceedings of the 2018 Conference on Empirical Methods in Natural Language Processing",
    month = oct # "-" # nov,
    year = "2018",
    address = "Brussels, Belgium",
    publisher = "Association for Computational Linguistics",
    url = "https://aclanthology.org/D18-1222/",
    doi = "10.18653/v1/D18-1222",
    pages = "1971--1979"
}

@article{Dadalto2024EvidenceOL,
  title={Evidence of large-scale conceptual disarray in multi-level taxonomies in Wikidata},
  author={At{\'i}lio A. Dadalto and Jo{\~a}o Paulo A. Almeida and Claudenir M. Fonseca and Giancarlo Guizzardi},
  journal={Semantic Web},
  year={2024},
  url={https://api.semanticscholar.org/CorpusID:268300153}
}

@article{Ferranti2024FormalizingAV,
  title={Formalizing and validating Wikidata’s property constraints using SHACL and SPARQL},
  author={Nicolas Ferranti and Jairo Francisco de Souza and Shqiponja Ahmetaj and Axel Polleres},
  journal={Semantic Web},
  year={2024},
  url={https://api.semanticscholar.org/CorpusID:271943502}
}

@InProceedings{low-redun,
author="Li, Wentao
and Zhou, Huachi
and Dong, Junnan
and Zhang, Qinggang
and Li, Qing
and Baciu, George
and Cao, Jiannong
and Huang, Xiao",
editor="Gonz{\'a}lez-Gonz{\'a}lez, Carina S.
and Fern{\'a}ndez-Manj{\'o}n, Baltasar
and Li, Frederick
and Garc{\'i}a-Pe{\~{n}}alvo, Francisco Jos{\'e}
and Sciarrone, Filippo
and Spaniol, Marc
and Garc{\'i}a-Holgado, Alicia
and Area-Moreira, Manuel
and Hemmje, Matthias
and Hao, Tianyong",
title="Constructing Low-Redundant and High-Accuracy Knowledge Graphs for Education",
booktitle="Learning Technologies and Systems",
year="2023",
publisher="Springer International Publishing",
address="Cham",
pages="148--160",
isbn="978-3-031-33023-0"
}

@inproceedings{Ferrarotti2009FirstOrderTA,
  title={First-Order Types and Redundant Relations in Relational Databases},
  author={Flavio Ferrarotti and Alejandra Lorena Paoletti and Jos{\'e} Maria Turull Torres},
  booktitle={ER Workshops},
  year={2009},
  url={https://api.semanticscholar.org/CorpusID:6537462}
}

@inproceedings{capobianco2020ontodrift,
  title={Ontodrift: a semantic drift gauge for ontology evolution monitoring},
  author={Capobianco, Giuseppe and Cavaliere, Danilo and Senatore, Sabrina and others},
  booktitle={CEUR Workshop Proceedings},
  volume={2821},
  pages={1--10},
  year={2020},
  organization={CEUR-WS}
}

@article{STAVROPOULOS201987,
title = {SemaDrift: A hybrid method and visual tools to measure semantic drift in ontologies},
journal = {Journal of Web Semantics},
volume = {54},
pages = {87-106},
year = {2019},
note = {Managing the Evolution and Preservation of the Data Web},
issn = {1570-8268},
doi = {https://doi.org/10.1016/j.websem.2018.05.001},
url = {https://www.sciencedirect.com/science/article/pii/S1570826818300258},
author = {T.G. Stavropoulos and S. Andreadis and E. Kontopoulos and I. Kompatsiaris}
}

@misc{yao2019kgbertbertknowledgegraph,
      title={KG-BERT: BERT for Knowledge Graph Completion}, 
      author={Liang Yao and Chengsheng Mao and Yuan Luo},
      year={2019},
      eprint={1909.03193},
      archivePrefix={arXiv},
      primaryClass={cs.CL},
      url={https://arxiv.org/abs/1909.03193}, 
}

@inproceedings{Yan2020LearningCK,
    title = "Learning Contextualized Knowledge Structures for Commonsense Reasoning",
    author = "Yan, Jun  and
      Raman, Mrigank  and
      Chan, Aaron  and
      Zhang, Tianyu  and
      Rossi, Ryan  and
      Zhao, Handong  and
      Kim, Sungchul  and
      Lipka, Nedim  and
      Ren, Xiang",
    editor = "Zong, Chengqing  and
      Xia, Fei  and
      Li, Wenjie  and
      Navigli, Roberto",
    booktitle = "Findings of the Association for Computational Linguistics: ACL-IJCNLP 2021",
    month = aug,
    year = "2021",
    address = "Online",
    publisher = "Association for Computational Linguistics",
    url = "https://aclanthology.org/2021.findings-acl.354/",
    doi = "10.18653/v1/2021.findings-acl.354",
    pages = "4038--4051"
}

@ARTICLE{IS-A,
  author={Brachman},
  journal={Computer}, 
  title={What IS-A Is and Isn't: An Analysis of Taxonomic Links in Semantic Networks}, 
  year={1983},
  volume={16},
  number={10},
  pages={30-36},
  doi={10.1109/MC.1983.1654194}}

@ARTICLE{Wang2023RiverON,
author={Wang, Kai and Lin, Dan and Luo, Siqiang},
journal={ IEEE Transactions on Knowledge \& Data Engineering },
title={{ Graph Percolation Embeddings for Efficient Knowledge Graph Inductive Reasoning }},
year={2025},
volume={37},
number={03},
ISSN={1558-2191},
pages={1198-1212},
doi={10.1109/TKDE.2024.3508064},
url = {https://doi.ieeecomputersociety.org/10.1109/TKDE.2024.3508064},
publisher={IEEE Computer Society},
address={Los Alamitos, CA, USA},
month=mar}

@misc{patelschneider2024classorderdisorderwikidata,
      title={Class Order Disorder in Wikidata and First Fixes}, 
      author={Peter F. Patel-Schneider and Ege Atacan Doğan},
      year={2024},
      eprint={2411.15550},
      archivePrefix={arXiv},
      primaryClass={cs.IR},
      url={https://arxiv.org/abs/2411.15550}, 
}

@article{SHENOY2022100679,
title = {A study of the quality of Wikidata},
journal = {Journal of Web Semantics},
volume = {72},
pages = {100679},
year = {2022},
issn = {1570-8268},
doi = {https://doi.org/10.1016/j.websem.2021.100679},
url = {https://www.sciencedirect.com/science/article/pii/S1570826821000536},
author = {Kartik Shenoy and Filip Ilievski and Daniel Garijo and Daniel Schwabe and Pedro Szekely}
}

@article{Wikidata,
author = {Vrande\v{c}i\'{c}, Denny and Kr\"{o}tzsch, Markus},
title = {Wikidata: a free collaborative knowledgebase},
year = {2014},
issue_date = {October 2014},
publisher = {Association for Computing Machinery},
address = {New York, NY, USA},
volume = {57},
number = {10},
issn = {0001-0782},
url = {https://doi.org/10.1145/2629489},
doi = {10.1145/2629489},
journal = {Commun. ACM},
month = sep,
pages = {78–85},
numpages = {8}
}

\end{document}